\documentclass[letterpaper, 10 pt, conference]{ieeeconf}  

\usepackage{hyperref}

\IEEEoverridecommandlockouts                              

\usepackage{multirow}
\usepackage{biblatex}

\title{\LARGE \bf
Skill2vec: Machine Learning Approach for Determining the Relevant Skills from Job Description
}

\author{
  Le Van-Duyet\\
  \texttt{me@duyet.net}
  \and
  Vo Minh Quan\\
  \texttt{95.vominhquan@gmail.com}
  \and
  Dang Quang An\\
  \texttt{an.dang1390@gmail.com}
}

\begin{document}

\maketitle
\thispagestyle{empty}
\pagestyle{empty}

\begin{abstract}

Unsupervise learned word embeddings have seen tremendous success in numerous Natural Language Processing (NLP) tasks in recent years. The main contribution of this paper is to develop a technique called Skill2vec, which applies machine learning techniques in recruitment to enhance the search strategy to find candidates possessing the appropriate skills. Skill2vec is a neural network architecture inspired by Word2vec, developed by Mikolov et al. in 2013. It transforms skills to new vector space, which has the characteristics of calculation and presents skills relationships.
We conducted an experiment evaluation manually by a recruitment company's domain experts to demonstrate the effectiveness of our approach.

\end{abstract}
\section{INTRODUCTION}

Recruiters in information technology domain have met the problem finding appropriate candidates by their skills. In the resume, the candidate may describe one skill in different ways or skills could be replaced by others. The recruiters may not have the domain knowledge to know if one's skills are fit or not, so they can only find ones with matched skills.

In order to cope with the problem, one should try to find the relatedness of skills. There are some approaches: building a dictionary manually, ontology approach, natural language processing methods, etc. In this study, we apply a word embedding method Word2Vec, using skills from online job post descriptions. We treat skills as terms, job posts as documents and find the relatedness of these skills.

\section{RELATED WORK}

To find relatedness of skills, Simon Hughes \cite{simon2015how} from Dice proposed an approach using Latent Semantic Analysis with an assumption that skills are related to skills which occur in the same context, and here contexts are job posts. This approach will build a term-document matrix and use Singular Value Decomposition to reduce the dimensionality. The cons of this approach is that when we have new data coming, we can not update the old term-document matrix, this leads to difficulties in maintaining the model, as the change of trend in this domain is high.

Google's Data Scientists also face the same problems in Cloud Jobs API \cite{posse2016cloud}. Their solution is to build a skill ontology defining around 50,000 skills in 30 job domain with relationships such as  is\_a, related\_to, etc. This approach can represent complicate relationships between skills and jobs, but building such an ontology costs so much time and effort.

To overcome the problem of relevant term, \cite{huang2003relevant} present a new, effective log-based approach to relevant term extraction and term suggestion.

The goal of \cite{bhat2004finding} is to develop an automated system that discovers additional names for an entity given just one of its names, using Latent semantic analysis (LSA) \cite{berry1996low}. In the example of authors, the city in India referred to as Bombay in some documents may be referred to as Mumbai in others because its name officially changed from the former to the latter in 1995.

\cite{lau2002introducing} is the introduction of an ontology-based skills management system at Swiss Life and the lessons learned from the utilisation of the methodology, present a methodology for application-driven development of ontologies that is instantiated by a case study.

\section{WORD2VEC ARCHITECTURE}

Word2Vec is a group of models proposed by Mikolov et al in 2013 \cite{journals/corr/abs-1301-3781}. It consists of 2 models: continuous bag-of-words and continuous Skip-gram, both are shallow neural networks that try to learn distributed representations of words with the target is to maximize the accuracy while minimizing the computational complexity. In the continuous bag-of-words architecture, the model predicts the current word from a window of surrounding context words. On the other hand, Skip-gram model try to predict surrounding context words based on the current word. In this work, we focus on Skip-gram model as it is known to be better with infrequent words and it also give slightly better result in our experiment.

The model consists of three layers: input layer, one hidden layer and output layer. The input layer take a word encoded using 1-of-V encoding (also known as one-hot encoding), where V is size of the dictionary. The word is then fed through the linear hidden layer to the output layer, which is a softmax classifier. The output layer will try to predict words within window size before and after current word. Using stochastic gradient descent and back propagation, we train the model until it converges.

This model takes vector dimensionality and window size as parameters. The author found that increasing the window size improves the quality of the word vector, and yet it increases the computational complexity.

\section{Methodology}

\subsection{Data collecting and processing}

Choosing a universal data set for the model is extremely important, the data should be large enough and should have balanced distributions over words (i.e. skills).

\begin{figure}[!htbp]
\centering
\begin{tikzpicture}[node distance=2.5cm]
                    
    \node (linkedin) [io, text width=2cm] {Career websites};
    \node (skill_standard) [process, right of=linkedin, xshift=2.5cm] {(1) Standard skills};
    \draw [arrow] (linkedin) -- node[anchor=south] {collect} node[anchor=north] {cleaning} (skill_standard);

    \node (dice) [io, below of=linkedin] {Career websites};
    \node (jd_raw) [process, below of=skill_standard] {Job descriptions};

    \draw [arrow] (dice) -- node[anchor=south] {collect} (jd_raw);
    
    \draw [arrow,dashed] (skill_standard) -- node[anchor=east] {(*)} node[anchor=west] {parser} (jd_raw);

    \node (training_data) [io, below of=jd_raw] {(2) Training data};
    \draw [arrow] (jd_raw) -- (training_data);
\end{tikzpicture}
\caption{Pipeline of data collecting and processing}
\label{fig:data-collect-pipeline}
\end{figure}
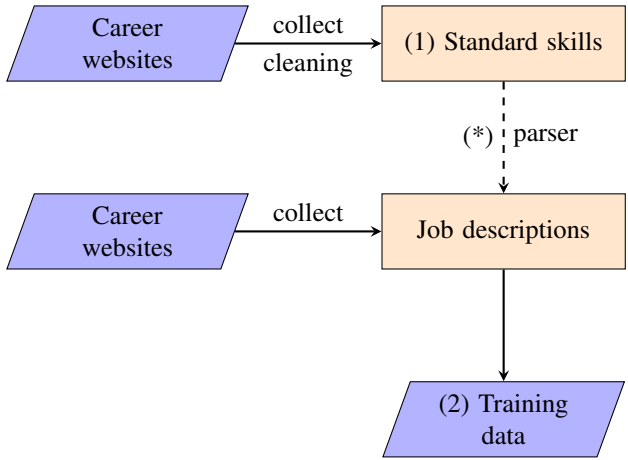

There are two dataset we need to concern, one (1) is the standard skills dictionary for the parser and another (2) is skills for training model; follow the figure \ref{fig:data-collect-pipeline}.

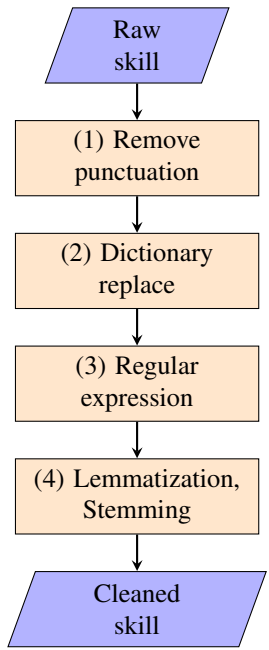
\begin{figure}[!htbp]
\centering
\begin{tikzpicture}[node distance=1.5cm]
                    
    \node (raw_data) [io, text width=1.3cm] {Raw skill};
    \node (step_1) [process, below of=raw_data] {(1) Remove punctuation};
    \node (step_2) [process, below of=step_1] {(2) Dictionary replace};
    \node (step_3) [process, below of=step_2] {(3) Regular expression};
    \node (step_4) [process, below of=step_3] {(4) Lemmatization, Stemming};
    \node (clean_data) [io, below of=step_4, text width=1.8cm, inner sep=0pt] {Cleaned skill};
    
    \draw [arrow] (raw_data) -- (step_1);
    \draw [arrow] (step_1) -- (step_2);
    \draw [arrow] (step_2) -- (step_3);
    \draw [arrow] (step_3) -- (step_4);
    \draw [arrow] (step_4) -- (clean_data);
    
\end{tikzpicture}
\caption{Pipeline ò cleaning skills}
\label{fig:data-cleaning-pipeline}
\end{figure}

First, we collected and prepared a large dictionary of skills. With this dictionary, we can extract a set of skills from raw job descriptions. Skillss need to be cleaned into unique skills because there are many way to present a skill in job description (\textit{i.e. OOP or Object-oriented programming}). Figure~\ref{fig:data-cleaning-pipeline} briefly depicts the concept of the cleaning process.

After that, we had the dictionary of skills ready for parsing. We collected a huge number of job descriptions from Dice.com - one of the most popular career website about Tech jobs in USA. From these job descriptions, we extract skills for each one by using our skills dictionary (1). Now, the dataset is presented by a list of collections of skills based on job descriptions. After crawling, we got a total of 5GB with more than 1,400,000 job descriptions.
From these data, we extracted skills and performed as a list of skills in the same context, the context here includes skills in the same job description.
The dataset is published at \href{https://github.com/duyetdev/skill2vec-dataset}{https://github.com/duyetdev/skill2vec-dataset}\\

The data structure is shown in table \ref{tab:data_structure}.

\begin{table}[h]
    \def\arraystretch{1.5}
    \centering
    \caption{Data structure}
    \label{tab:data_structure}
    \begin{tabular}{|c|c|}
        \hline
        Job description & Context skills \\ \hline 
        JD1 & Java, Spark, Hadoop, Python  \\ \hline 
        JD2 & Python, Hive \\ \hline
        JD3 & Python, Flask, SQL \\ \hline
        $\cdots$ & $\cdots$ \\ \hline 
    \end{tabular}

\end{table}

\subsection{Skill2vec architecture}
\begin{figure}[h]
\centering
\includegraphics[width=0.5\textwidth]{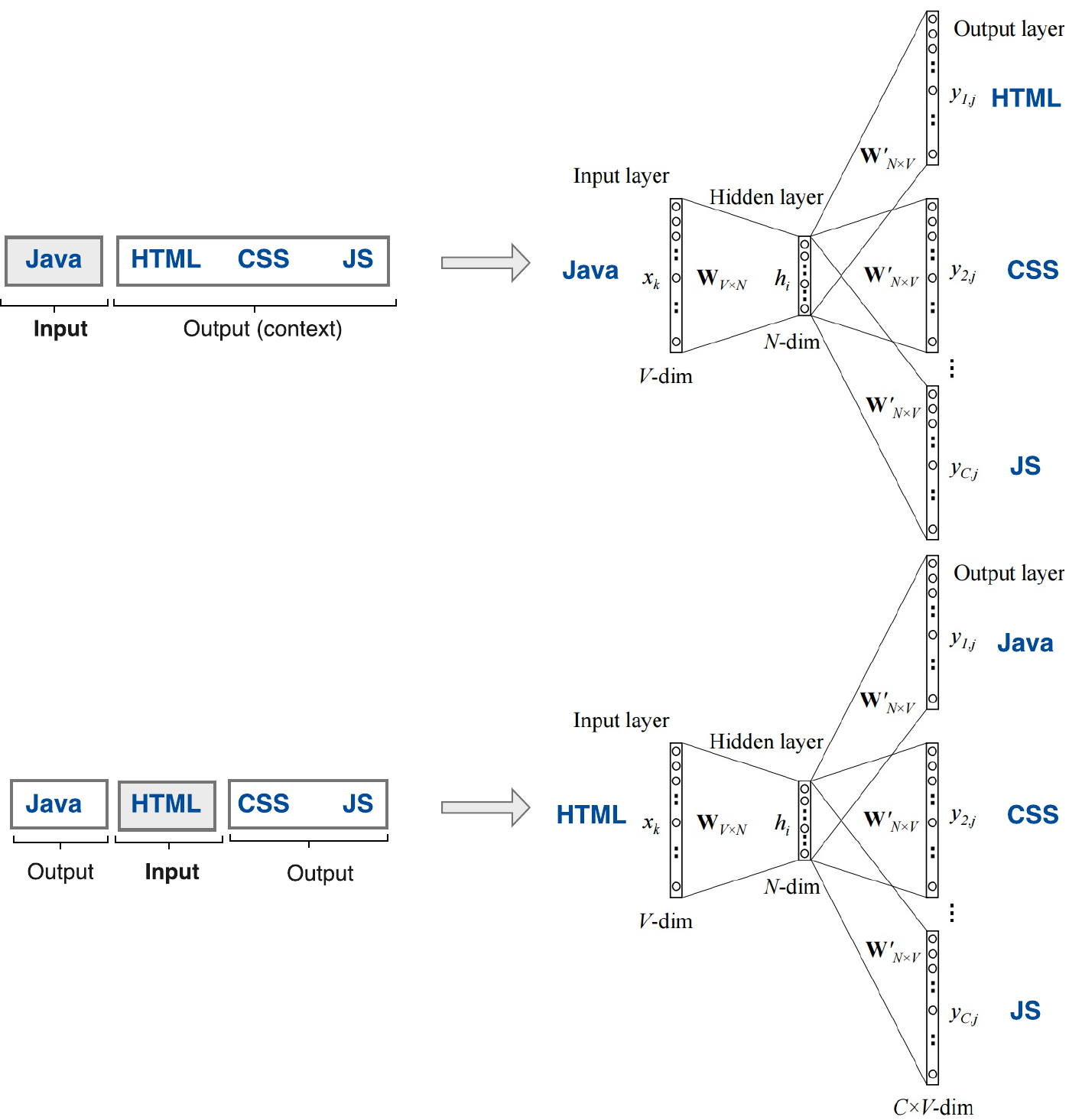}
\caption{Skill2vec architecture}
\label{fig:skill2vec_how_to_make_training_data}
\end{figure}

In this paper, for training the dataset, we used a neural network inspired by Word2Vec model as mentioned above. Here we treated our skills as words in Word2Vec model. In this study, with the documents contain only the skills, we chose the maximum window size, implied that every skills in the same job description are related to each other. For the vector dimensions, after some point, adding more dimensions provides diminishing improvements, so we chose this parameter empirically. To honour the work of Word2Vec model as it holds a big part in our study, we name our model Skill2Vec. Figure \ref{fig:skill2vec_how_to_make_training_data} briefly describes our Skill2Vec model.

\section{EXPERIMENTAL SETUP}
To evaluate our method, we have an expert team assesses the result following these steps:
\begin{enumerate}
\item Pick 200 skills randomly from our dictionary.
\item Our system will return top 5 "nearest" skills for each.
\item The expert team will check if these top 5 "nearest" skills are relevant or not.
\end{enumerate}

The experiment showed that 78\% of skills returned by our model is truly relevant to the input skill. We present the experimental results in table \ref{tab:results-top-5}

\begin{table}[h]
\centering
\def\arraystretch{1.5}
\caption{Query top 5 relevant skills}
\label{tab:results-top-5}
\begin{tabular}{|c|c|}
\hline
Query skill             & Top relevant skills \\ \hline

\multirow{5}{*}{HTML5}   & css3                 \\ \cline{2-2} 
                         & bootstrap            \\ \cline{2-2} 
                         & front\_end           \\ \cline{2-2} 
                         & angular              \\ \cline{2-2} 
                         & responsive           \\ \hline
\multirow{5}{*}{OOP}     & OOD                  \\ \cline{2-2} 
                         & Objective            \\ \cline{2-2} 
                         & Java                 \\ \cline{2-2} 
                         & Multithread          \\ \cline{2-2} 
                         & Software Debug       \\ \hline

\multirow{5}{*}{Hadoop} & Pig                 \\ \cline{2-2} 
                        & Hive                \\ \cline{2-2} 
                        & HBase               \\ \cline{2-2} 
                        & Big Data            \\ \cline{2-2} 
                        & Spark               \\ \hline
\multirow{5}{*}{Scala}  & Zookeeper           \\ \cline{2-2} 
                        & Spark               \\ \cline{2-2} 
                        & Data System         \\ \cline{2-2} 
                        & Sqoop               \\ \cline{2-2} 
                        & solrcloud           \\ \hline
\multirow{5}{*}{Hive}   & Pig                 \\ \cline{2-2} 
                        & HDFS                \\ \cline{2-2} 
                        & Hadoop              \\ \cline{2-2} 
                        & Spark               \\ \cline{2-2} 
                        & Impala              \\ \hline
\end{tabular}
\end{table}

\section{CONCLUSION}
In this paper, we developed a relationship network between skills in recruitment domain by using the neural net
inspired by Word2vec model. We observed that it is possible to train high quality word vectors using very simple
model architectures due to lower cost of computation. Moreover, it is possible to compute very accurate high
dimensional word vectors from a much larger dataset. Using Skip-gram architecture and an advanced technique for
preprocessing data, the result seems to be impressive. The result of our work can contribute to building the matching
system between candidates and job post. In the other hand, candidates can find the gap between the job post
requirements and their ability, so they can find the suitable trainings.

A direction we can follow in the future: adding domain in training model, for example: Between \textit{Python}, \textit{Java}, and \textit{R}, in \textit{Data Science} domain, \textit{Python} and \textit{R} are more relevant than \textit{Java}, however in \textit{Back End} domain, \textit{Python} and \textit{Java} are more relevant than \textit{R}.


\printbibliography

\end{document}